%% file: icml2025.tex
\documentclass{article}

\usepackage{microtype}
\usepackage{graphicx}
\usepackage{subfigure}
\usepackage{booktabs} 
\usepackage{enumitem}
\usepackage{hyperref}

\usepackage[accepted]{icml2025}

\usepackage{amsmath}
\usepackage{amssymb}
\usepackage{mathtools}
\usepackage{amsthm}
\usepackage[most]{tcolorbox}

\usepackage{makecell}
\usepackage{multirow}
\usepackage{bbding}
\usepackage{colortbl} 
\usepackage{xcolor}  
\usepackage{float}
\usepackage{CJKutf8}

\usepackage[capitalize,noabbrev]{cleveref}

\theoremstyle{plain}

\theoremstyle{definition}

\theoremstyle{remark}

\usepackage[textsize=tiny]{todonotes}

\icmltitlerunning{A Rigorous Examination of Mitigation Strategies for LLM Benchmark Data Contamination}

\begin{document}

\twocolumn[
\icmltitle{The Emperor's New Clothes in Benchmarking? A Rigorous Examination of Mitigation Strategies for LLM Benchmark Data Contamination}



\icmlsetsymbol{equal}{*}

\begin{icmlauthorlist}
\icmlauthor{Yifan Sun}{equal,yyy}
\icmlauthor{Han Wang}{equal,yyy}
\icmlauthor{Dongbai Li}{equal,yyy}
\icmlauthor{Gang Wang}{yyy}
\icmlauthor{Huan Zhang}{yyy}
\end{icmlauthorlist}

\icmlaffiliation{yyy}{University of Illinois Urbana-Champaign}

\icmlcorrespondingauthor{Yifan Sun}{yifan50@illinois.edu}

\icmlkeywords{Machine Learning, ICML}

\vskip 0.3in
]



%
\printAffiliationsAndNotice{\icmlEqualContribution} 

\input{sections/0_abstract}
\input{sections/1_intro}
\input{sections/2_related_work}
\input{sections/3_Method}

\input{sections/4_Evaluation_pipeline}

\input{sections/5_Results}

\input{sections/6_conclusion}

\nocite{langley00}

\bibliography{icml2025}
\bibliographystyle{icml2025}

\input{sections/7_Appendix}

\end{document}

%% file: sections/0_abstract.tex
\begin{abstract}
Benchmark Data Contamination (BDC)—the inclusion of benchmark testing samples in the training set—has raised increasing concerns in Large Language Model (LLM) evaluation, leading to falsely inflated performance estimates and undermining evaluation reliability. To address this, researchers have proposed various mitigation strategies to update existing benchmarks, including modifying original questions or generating new ones based on them. However, a rigorous examination of the effectiveness of these mitigation strategies remains lacking. In this paper, we design a systematic and controlled pipeline along with two novel metrics—\textit{fidelity} and \textit{contamination resistance}—to provide a fine-grained and comprehensive assessment of existing BDC mitigation strategies. Previous assessment methods, such as accuracy drop and accuracy matching, focus solely on aggregate accuracy, often leading to incomplete or misleading conclusions. Our metrics address this limitation by emphasizing \textit{question-level} evaluation result matching. Extensive experiments with 10 LLMs, 5 benchmarks, 20 BDC mitigation strategies, and 2 contamination scenarios reveal that no existing strategy significantly improves resistance over the vanilla case (\textit{i.e.}, no benchmark update) across \textit{all} benchmarks, and none effectively balances fidelity and contamination resistance. These findings underscore the urgent need for designing more effective BDC mitigation strategies. Our code repository is available at \href{https://github.com/ASTRAL-Group/BDC_mitigation_assessment}{https://github.com/ASTRAL-Group/BDC\_mitigation\_assessment}.
\end{abstract}

%% file: sections/1_intro.tex
\section{Introduction}
\label{sec:intro}

Benchmarking Large Language Models (LLMs) has recently become a critical area of focus \citep{white2024livebench,xia2024top, guha2024legalbench, zeng2024air, lin2024wildbench,ni2024mixeval}, driven by the rapid increase in their number and capacity \citep{achiam2023gpt, dubey2024llama, team2024qwen2, team2023gemini, team2024gemma}. Reliable and high-quality evaluation benchmarks are essential to provide comprehensive and accurate assessments of LLM capabilities. However, as modern LLMs are trained on vast amounts of web-scraped data, concerns have emerged regarding benchmark samples inadvertently appearing in their training sets. Consequently, it is challenging to determine whether the model just simply memorizes answers to difficult test questions to achieve a better performance~\cite{oren2023proving, zhu2024inference}. This phenomena, known as \textbf{Benchmark Data Contamination (BDC)}, results in falsely inflated performance metrics, thereby undermining the reliability of evaluation conclusions \citep{zhou2023don, jiang2024does, sainz2023nlp}.

\begin{figure}[!htbp]
    \centering
    \includegraphics[width=\linewidth]{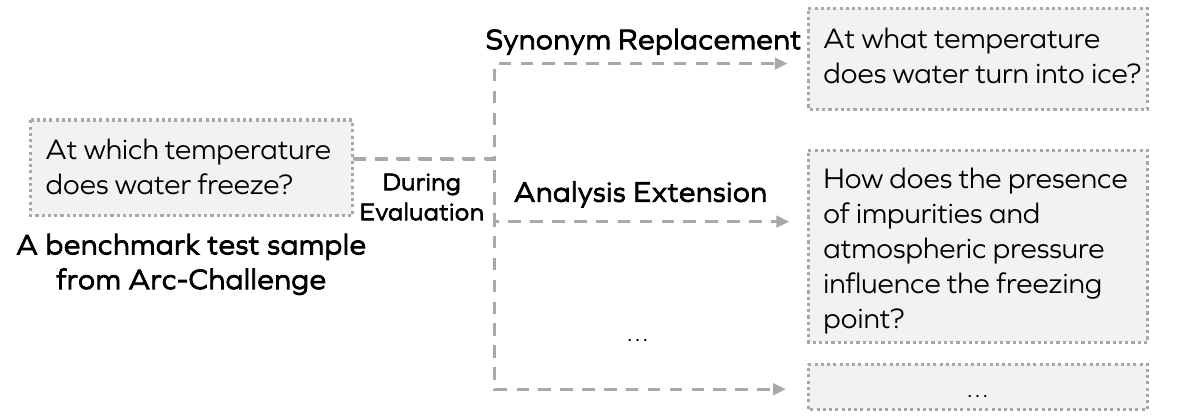}
    \vspace{-0.5cm}
    \caption{\textbf{Illustration of BDC mitigation strategies.} BDC mitigation strategies, such as synonym replacement and analysis extension \citep{ying2024automating}, 
    \textit{update} benchmark questions to reduce the risk of direct memorization.}
\end{figure}


\begin{figure*}[ht]
    \centering
    \includegraphics[width=\textwidth]{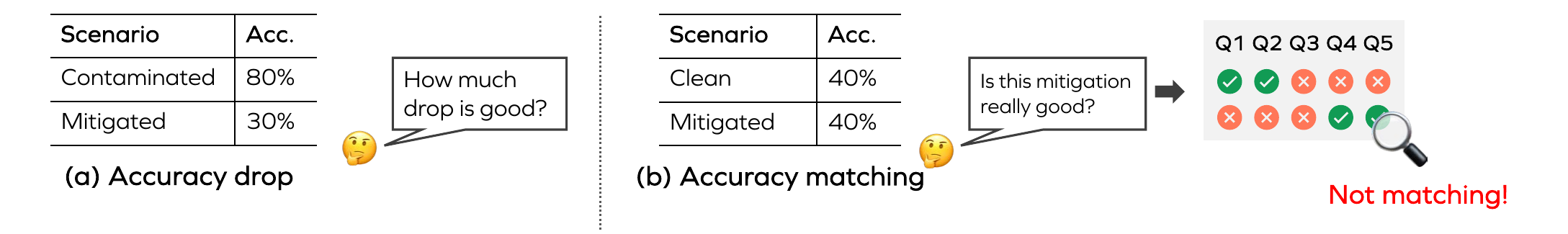}
    \vspace{-20pt}
    \caption{\textbf{The limitations of existing approaches for assessing BDC mitigation strategies}: 
    (a) \textbf{Accuracy drop} measures the performance decline between contaminated accuracy and mitigated accuracy, but does not account for the clean accuracy, making it unclear how much drop indicates effective mitigation.
    (b) \textbf{Accuracy matching} requires that the mitigated accuracy restores clean accuracy. However, as shown in the example, even when the accuracies match, the \textit{question-level} evaluation results differ significantly (\textit{e.g.}, correctly answering the 1st and 2nd questions versus the 4th and 5th). This discrepancy suggests that the updated benchmark may evaluate different aspects of model capacity compared to the original benchmark. As a result, the mitigation strategy may fail to preserve the original benchmark’s evaluation objective and could be ineffective.}
    \label{fig:previous_methods}
    \vspace{-8pt}
\end{figure*}
\vspace{-1cm}
To mitigate BDC, creating new benchmark datasets from scratch is a potential solution, but this process is often prohibitively expensive and labor-intensive\footnote{For instance, curating the GPQA dataset \citep{rein2023gpqa}, which contains 448 multiple-choice questions written by domain experts, required over \$120,000 \citep{rein2024}. Similarly, the recently introduced HLE benchmark \citep{phan2025humanitysexam} has allocated \$500,000 to collect high-quality benchmark questions.}. Moreover, some existing benchmark datasets, such as MMLU \citep{hendrycks2020measuring} and GSM8K \citep{cobbe2021training}, are already of high quality and accurately reflect real-world question distributions within their respective domains. Rather than retiring such well-established benchmarks, ongoing efforts
\begin{table}[H]
\centering
\caption{Definition of different evaluation scenarios based on the contamination status of the LLM and the benchmark version used.}
\resizebox{0.7\linewidth}{!}{%
\begin{tabular}{c|cc}
\toprule[1.5pt]
\textbf{Scenario}     & \textbf{LLM}          & \textbf{Benchmark} \\ \hline
Clean                 & Uncontaminated       & Original           \\ \hline
Contaminated          & Contaminated           & Original            \\ \hline
Mitigated             & Contaminated         & Updated            \\ \bottomrule[1.5pt]
\end{tabular}%
}
\vspace{-8pt}
\label{tab:evaluation-scenarios}
\end{table}
aim to update them or generate new questions based on these benchmarks to \textbf{mitigate} BDC \citep{zhu2023clean,zhu2024dynamic,zhu2024inference, ying2024automating}. For example, a straightforward approach is to paraphrase original questions, reducing the risk of models naively leveraging memorized answers.

\begin{figure*}[ht]
    \centering
    \includegraphics[width=\textwidth]{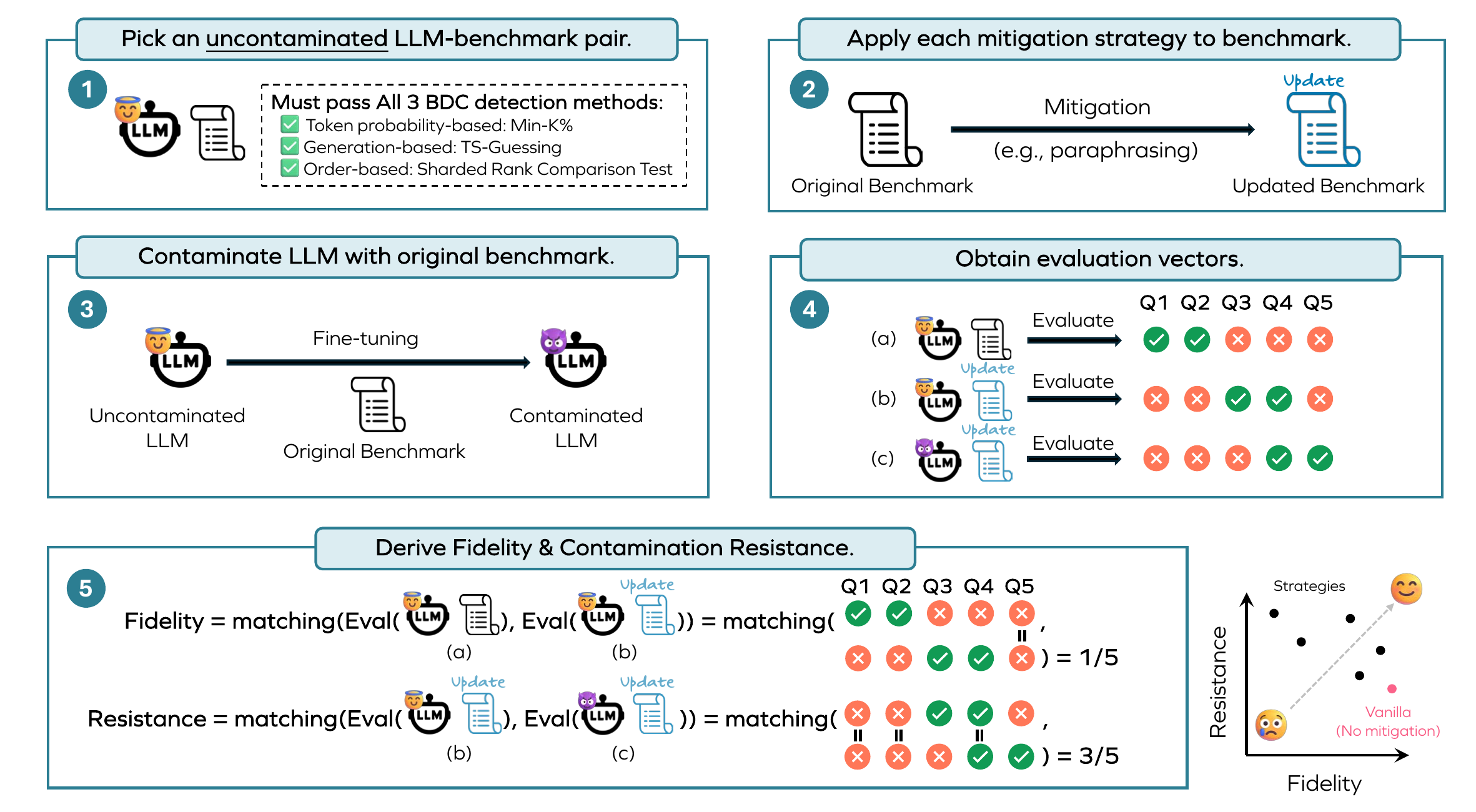}
    \vspace{-22pt}
    \caption{\textbf{Overview of our pipeline for assessing BDC mitigation strategies}: (1) We select an LLM-benchmark pair and ensure it passes three BDC detection methods to confirm it is uncontaminated, a crucial step for reliable ``clean" evaluation results (\textsection\ref{sec:pipeline_filtering}). (2) Each mitigation strategy is applied separately to the original benchmark to produce an updated benchmark; 20 strategies are examined in total (\textsection\ref{sec:pipeline_mitigation}). (3) The uncontaminated LLM is fine-tuned on the original benchmark dataset. Two contamination recipes (mild and intensive) are tested to ensure robust conclusions and three validation checks are performed to confirm the effectiveness of the contamination process (\textsection\ref{sec:pipeline_contamination}). (4) Evaluation vectors are computed for: (a) uncontaminated LLM with the original benchmark, (b) uncontaminated LLM with the updated benchmark, and (c) contaminated LLM with the updated benchmark (\textsection\ref{sec:pipeline_obtain}). (5) Fidelity and resistance are derived based on the degree of matching between these evaluation vectors (\textsection\ref{sec:method}). An effective mitigation strategy should achieve high scores in both metrics.
    }
    \label{fig:pipeline}
    \vspace{-10pt}
\end{figure*}

\begin{tcolorbox}[ colback=white,  colframe=red!70!black, title=Our Research Question ] Each BDC \textit{mitigation} strategy yields an \textit{updated} benchmark. We focus on a thorough and rigorous examination towards the effectiveness of different BDC mitigation strategies. \end{tcolorbox} 




However, it is crucial to assess the effectiveness of different BDC mitigation strategies systematically. For example, whether surface paraphrasing can indeed alleviate the effects of BDC is under question. Nevertheless, current practices for assessing BDC mitigation strategies have clear limitations, as illustrated in Fig.~\ref{fig:previous_methods}: (a) \textbf{Accuracy drop.} Some previous studies regard a mitigation strategy as successful if the contaminated LLM’s accuracy on the updated benchmark (\textit{i.e.,} mitigated accuracy) is lower than its accuracy on the original benchmark (\textit{i.e.,} contaminated accuracy)~\citep{zhu2024dynamic}. However, without referencing the model’s performance on the original benchmark before any contamination (\textit{i.e.,} clean accuracy), it is unclear how much of a drop is meaningful. (b) \textbf{Accuracy matching.} Other works assess mitigation strategies by comparing clean accuracy with mitigated accuracy, expecting them to match~\citep{zhu2023clean,zhu2024inference,ying2024automating}. Yet, accuracy is only an aggregate metric. Focusing solely on matching the \textit{scalar} accuracy is not sufficient and can even be misleading. For example, in the case shown in Fig. \ref{fig:previous_methods}(b), even if scalar accuracy aligns, the strategy fails to recover the clean \textbf{question-wise} evaluation results. Consequently, the mitigation strategy may alter the original benchmark’s evaluation objective, putting its effectiveness into question.


In this paper, we present a comprehensive and rigorous framework for assessing BDC mitigation strategies (Fig. \ref{fig:pipeline}). We identify two key desiderata for an effective strategy:
\textbf{(1) Fidelity:} For a high-fidelity strategy, if the clean LLM answers the original question correctly, it also answers the updated question correctly; if it fails on the original question, it also fails on the updated version.
\textbf{(2) Contamination Resistance:} For a contamination-resistant strategy, even if the LLM has been contaminated by the original dataset, its ability to answer each question in the updated benchmark remains unchanged.

By employing the normalized Hamming distance and jointly evaluating these metrics, our framework emphasizes \textbf{question-wise matching}, offering a fine-grained and multi-faceted assessment of mitigation strategies. 

\textbf{Our main contributions are summarized as follows}:
\begin{itemize}
    \item We identify the limitations of existing approaches for assessing BDC mitigation strategies and propose two novel metrics, fidelity and contamination resistance~(\textsection\ref{sec:method}).
    \item We design a scientific and controlled pipeline to assess BDC mitigation strategies. Different from previous studies, extensive checks are performed to confirm that each LLM-benchmark pair is uncontaminated prior to manual contamination, ensuring the validity of clean evaluation results.
    Two contamination recipes that simulate real-world data contamination scenarios are examined~(\textsection\ref{sec:pipeline}).
    \item Through experiments with 10 LLMs, 5 benchmarks and 20 BDC mitigation strategies, we find that none of the existing mitigation strategies offers statistically significantly higher resistance than the vanilla approach (\textit{i.e.,} no dataset update) across \textit{all} benchmarks. More critically, none achieves strong fidelity and contamination resistance simultaneously, highlighting the need for designing more effective mitigation strategies~(\textsection\ref{sec:results}). 
\end{itemize}

%% file: sections/2_related_work.tex
\section{Related Work}




Existing research seeks to mitigate the impact of BDC through two primary strategies: curating new benchmarks and updating existing benchmarks \citep{xu2024benchmark}. Recent works have proposed novel benchmarks to address contamination ~\cite{li2024latesteval, zhu2023dyval, jain2024livecodebench, li2024evocodebench, qian2024varbench, wu2024antileak}. While effective, this approach is costly and time-intensive, requiring significant human effort for labeling and maintenance.
An alternative strategy focuses on updating existing high-quality benchmarks, maximizing the utilization of well-established benchmarks while being more cost-effective and automated. Some methods modify evaluation samples while preserving their semantics ~\cite{zhu2024inference, zhu2024dynamic, zhu2023clean, li2024c, wang2021adversarial, xia2024top, haimes2024benchmark, zheng2024ali}. Others generate new samples with altered semantics based on original questions,  using advanced LLMs~\citep{ying2024automating}. However, in the latter case, the quality of generated samples is often limited by the task-specific capabilities of the underlying LLMs used in the generation process. We leave a discussion of BDC detection methods to Appendix \ref{Append:extended_related}.

%% file: sections/3_Method.tex
\section{Method}
\label{sec:method}

We focus exclusively on BDC mitigation strategies that update existing benchmarks, since introducing entirely novel ones can be difficult to automate and incurs high costs. Without a clear and thorough understanding of how well these mitigation strategies work, benchmark developers and evaluation practitioners risk making unnecessary changes to existing benchmarks that fail to actually reduce the impact of BDC. 
In this section, we propose two novel metrics to comprehensively assess BDC mitigation strategies.


\textbf{Notation and Setup.}
Let $\mathcal{M}$ be the space of LLMs, and let $\mathcal{D}$ be the space of datasets. 
Consider a benchmark dataset $D \in \mathcal{D}$ consisting of $n$ questions (\textit{e.g.}, multiple-choice questions), and let $M \in \mathcal{M}$ be an LLM that is not contaminated by $D$.
We define an \emph{evaluation function} 
\vspace{-2pt}
\begin{equation*}
R: \mathcal{M} \times \mathcal{D} \to \{0,1\}^n,
\end{equation*}
which takes as input an LLM-benchmark pair $(M, D)$ and outputs an \textbf{evaluation vector} in $\{0,1\}^n$. This evaluation vector is a critical component of our framework, as it captures the model's performance on the benchmark at a question-by-question level.
For each question $i \in \{1, \dots, n\}$, $R(M,D)_i=1$ indicates that $M$ answers the $i$-th question correctly, and $R(M,D)_i=0$ otherwise\footnote{In Appendix \ref{Append:continuous}, we discuss how our framework can be extended to cases where the evaluation scores are continuous.}. 


Let $M^D$ denote the version of $M$ that has been contaminated by $D$. Additionally, let $S$ represent a benchmark update strategy that transforms $D$ into $D^S$, with the goal of mitigating potential data contamination. 

\textbf{Metrics for Assessing the Mitigation Strategy. }
We propose the following criteria to assess $S$:

\textbf{(1) Fidelity:} Since the original benchmark is assumed to be of high quality, whether each question is answered correctly or incorrectly should reflect the model's true capabilities. For the updated benchmark, it is crucial that the clean model’s performance \textbf{on each question} aligns with its performance on the original benchmark. Specifically, if the clean model answers a question correctly (or incorrectly) in the original benchmark, it should also answer the corresponding updated question correctly (or incorrectly).  Formally, the evaluation vectors on $D$ and $D^S$ for the clean model $M$ should match:
    \begin{equation*}
        R(M, D) \approx R(M, D^S).
    \end{equation*}
    It is important to clarify why high fidelity is necessary. \textit{Low fidelity does not necessarily mean the updated benchmark is of poor quality}. Rather, it signals a significant deviation from the original benchmark. For example, consider a mathematical reasoning question where an aggressive rewording alters the problem's implicit assumptions, making it substantially easier or harder to solve. If the clean model originally answers the question correctly but fails after the benchmark update—or vice versa—it suggests that the update may have changed the problem's complexity or the aspect of model's capability being evaluated. As a result, a low fidelity score is assigned, and additional manual checks may be needed to ensure the factuality and quality of the updated benchmark. In such cases, such a strategy can no longer be considered as a fully {\em automated} mitigation strategy due to the need for manual post-hoc inspection. 

\textbf{(2) Contamination Resistance:} 
A contamination-resistant strategy ensures that an LLM does not gain any advantage on the \textit{updated} benchmark from being exposed to the \textit{original} benchmark. 
If the model was correct (or incorrect) on a question in the updated benchmark before contamination, it should remain correct (or incorrect) after contamination by the original benchmark. Formally, the evaluation vectors on $D^S$ should remain similar regardless of whether $M$ is contaminated by $D$ or not:
    \begin{equation*}
        R(M, D^S) \approx R(M^D, D^S).
    \end{equation*}
Note that we consider question-wise matching rather than just matching overall accuracy. Since $R(M,D)$, $R(M,D^S)$, and $R(M^D,D^S)$ are binary vectors, we use the \textit{normalized} Hamming distance \citep{hamming1950error}:
\begin{equation*}
\mathrm{H}(x, y) \;=\; \frac{1}{n} \sum_{i=1}^n \mathbf{1}\bigl[x_i \neq y_i\bigr].
\end{equation*}
With a benchmark dataset $D$ and a LLM $M$, we define the fidelity and resistance metrics for strategy $S$ as:
\begin{equation*}
\begin{aligned}
\mathrm{Fidelity}(S) &= 1 - \mathrm{H}\bigl(R(M,D),\; R(M,D^S)\bigr), \\
\mathrm{Resistance}(S) &= 1 - \mathrm{H}\bigl(R(M,D^S),\; R(M^D,D^S)\bigr).
\end{aligned}
\label{eqn}
\end{equation*}
\textbf{Discussion. }
We underline that an ideal benchmark update strategy must perform well in terms of both fidelity and resistance. If no update is performed (\emph{i.e., vanilla strategy}), fidelity is trivially $1$, but the resistance can be poor. On the other hand, if the original benchmark is replaced with something entirely unrelated (for example, turning GSM8K \citep{cobbe2021training} into a history-based benchmark), resistance may be high, and yet fidelity is lost. Hence, a solid approach should achieve high scores on both metrics.

%% file: sections/4_Evaluation_pipeline.tex
\section{Pipeline}
\label{sec:pipeline}
\subsection{Overview}

To compute fidelity and resistance metrics, it is essential to have access to both an uncontaminated LLM and its contaminated counterpart. However, obtaining both can be challenging in practice, especially when the contamination status of a given LLM is not transparent. To address this issue, we deliberately select \textit{uncontaminated} LLM-benchmark pairs and then \textit{manually contaminate} the LLMs. 

In this section, we present a carefully designed pipeline to systematically and thoroughly evaluate 20 existing BDC mitigation strategies. An overview of the pipeline is provided in Fig.~\ref{fig:pipeline}. Our framework incorporates two key improvements over existing approaches:  (1) thorough contamination checks to ensure the models are uncontaminated before manually introducing contamination, and (2) different contamination recipes to account for the diversity of real-world contamination scenarios. These components enable our controlled pipeline to yield solid, generalizable insights. 

In contrast, existing accuracy matching frameworks \citep{zhu2023clean,zhu2024inference,ying2024automating} fail to confirm that the LLM is uncontaminated before manual contamination. As a result, their claimed “clean” performance may be inaccurate, introducing noise into their conclusions. Additionally, these frameworks typically involve only one contamination recipe, weakening the robustness of their conclusions.




\subsection{LLM and Benchmark Selection}
\label{sec:pipeline_filtering}

\textbf{Benchmarks. }
We select five benchmarks for our primary experiments, four of which are commonly used in prior studies on BDC detection and mitigation~\citep{zhou2023don,shi2023detecting,zhu2023clean}: (1) Arc-Challenge (Arc-C)~\cite{clark2018think}, which focuses on grade-school science tasks; (2) MMLU~\cite{hendrycks2020measuring}, which evaluates comprehensive world knowledge; (3) TruthfulQA~\cite{lin2021truthfulqa}, which measures the truthfulness of LLM-generated answers; and (4) GSM8K~\cite{cobbe2021training}, which tests grade-school mathematics. We also include the recently released RepliQA~\cite{monteiro2024repliqa}, a question-answering benchmark with non-factual yet natural-looking contexts about fictional entities. Its recent release\footnote{This benchmark was released on December 9, 2024~\citep{monteiro2024repliqa}.} and non-factual nature ensure that none of the LLMs in our study have been contaminated by this benchmark, making it an ideal candidate for our controlled pipeline. Detailed benchmark information is provided in Appendix \ref{Append:model details}.

\textbf{LLMs. }
To ensure reliable conclusions free from potential noise, 
we make every effort to select LLMs uncontaminated prior to introducing manual contamination. To achieve this, we apply three BDC detection methods from distinct categories—Min-K\% Prob~\cite{shi2023detecting}, Sharded Rank Comparison Test~\cite{oren2023proving}, and TS-Guessing~\cite{deng2023investigating}—to 14 candidate models. We adopt a rigorous criterion: only models deemed uncontaminated by \textit{all} three detection methods on \textit{all} benchmarks are retained (see Appendix~\ref{Append:filtering} for detailed results). In the end, we select 10 popular LLMs, spanning parameter sizes from 3B to 34B and originating from different model publishers, ensuring a broad representation. Detailed model information is provided in Appendix \ref{Append:model details}.

\subsection{Mitigation Strategies}
\label{sec:pipeline_mitigation}

Our analysis focuses on BDC mitigation strategies that leverage existing benchmarks, categorized into two primary approaches: semantic-preserving and semantic-altering updates~\citep{xia2024top}. Within the semantic-preserving updates, we collect 11 distinct mitigation strategies: irrelevant context~\cite{wang2021adversarial}, relevant context~\cite{zhu2024dynamic}, syntactic modification~\cite{zhu2023clean, zhu2024inference, zhu2024dynamic}, synonym replacement~\cite{zhu2023clean, zhu2024inference, zhu2024dynamic}, typographical perturbation~\cite{wang2021adversarial}, translation (Chinese)~\cite{li2024c}, translation (French), back-translation~\cite{zhu2023clean}, choice paraphrasing~\cite{zhu2024dynamic}, additional incorrect choices~\cite{zhu2024dynamic}, and choice permutation~\cite{zhu2024dynamic}. These strategies can be systematically combined to create more complex ones. Our study encompasses both combinations proposed in prior work (\textit{i.e.}, Clean-Eval~\cite{zhu2023clean}, ITD~\cite{zhu2024inference}, and MPA~\cite{zhu2024dynamic}) and two new combinations introduced in this paper: MPA-Ques+Trans-CN and MPA-Choice+Trans-CN.
In addition to semantic-preserving strategies, we also examine semantic-altering strategies that generate evaluation samples with different semantics based on the original benchmark: mimicking, remember-understand extension, application extension, and analysis extension~\cite{ying2024automating}. In total, our study assesses 20 mitigation strategies, which, to the best of our knowledge, comprehensively cover all existing BDC mitigation strategies proposed to date. Detailed information is provided in Tab.~\ref{tab:mitigation}.


\begin{table*}[t!]
    \caption{\textbf{Overview of 20 BDC mitigation strategies assessed in our study.}
    The ``Scope'' column denotes the applicable objects of each mitigation strategy, categorized into Questions (Q) or Choices (C).
    }
\centering   
   \resizebox{0.98\linewidth}{!}{
   \begin{tabular}{ l  | l |  l }
   \toprule[1.5pt]
     \makecell[l]{\textbf{Mitigation Strategies}}  &  \textbf{Scope} & \textbf{Descriptions} \\ \midrule
     \multicolumn{3}{l}{\emph{Semantic-Preserving Updates (Single Strategy)}} \\
   $S_1$: Irrelevant Context & Q & Append irrelevant content (e.g., ``https://t.co/DlI9kw'') before the question \\ 
    $S_2$: Relevant Context & Q & Introduce a relevant scenario before the question \\ 
    $S_3$: Syntactic Modification & Q & Modify the syntactic structure of the question \\ 
     $S_4$: Synonym Replacement & Q & Replace certain words in the question with synonyms \\ 
     $S_5$: Typographical Perturbation & Q &  Introduce typos or minor spelling errors in the question \\
     $S_6$: Translation (Chinese) & Q \& C & Translate the question and choices into Chinese \\ 
     $S_7$: Translation (French)  & Q \& C & Translate the question and choices into French \\ 
     $S_8$: Back-translation & Q \& C & Translate the question and choices into Chinese and back to English \\
     $S_9$: Choice Paraphrasing  & C & Reword and restructure each choice \\       
     $S_{10}$: Additional Incorrect Choices & C & Add distractor choices \\ 
     $S_{11}$: Choices Permutation & C &  Rearrange the order of the choices \\ \midrule
     \multicolumn{3}{l}{\emph{Semantic-Preserving Updates (Combined Strategy)}} \\
     $S_{12}$: Clean-Eval & Q \& C & $S_3+S_4+S_8$ \\ 
     $S_{13}$: ITD & Q \& C &  $S_2+S_3+S_4+S_9$ \\ 
     \makecell[l]{$S_{14}$: MPA} & Q \& C &  \makecell[l]{$S_2+S_3+S_4+S_9+S_{10}+S_{11}$}
     \\  
      $S_{15}$: MPA-Ques + Trans-CN & Q \& C & $S_2+S_3+S_4+S_6$ \\ 
      $S_{16}$: MPA-Choice + Trans-CN & Q \& C & $S_6+S_9+S_{10}$ \\ \midrule
      \multicolumn{3}{l}{\emph{Semantic-Altering Updates}} \\
       \makecell[l]{$S_{17}$: Mimicking} & Q \& C & Generate samples with different concepts but similar styles  \\
       \makecell[l]{$S_{18}$: Remember-Understand Extension} & Q \& C & Generate samples that evaluate recall of facts and basic ideas \\
       \makecell[l]{$S_{19}$: Application  Extension} & Q \& C & Generate samples that require applying concepts to solve practical problems \\
       \makecell[l]{$S_{20}$: Analysis Extension} & Q \& C & Generate samples that evaluate the ability to analyze conceptual relationships \\
      
    \bottomrule[1.5pt]
    \end{tabular} }
    \vspace{-5pt}
    \vspace{-5pt}
    \label{tab:mitigation}
\end{table*}

\subsection{Model Contamination}
\label{sec:pipeline_contamination}

For each uncontaminated LLM-benchmark pair (10 $\times$ 5 = 50 pairs in total), we manually introduce contamination by full parameter fine-tuning the LLM on the benchmark dataset. To ensure a comprehensive assessment, we implement two distinct contamination recipes: (1) \textbf{Mild Contamination}: The benchmark data is mixed with 20,000 randomly selected samples from OpenOrca~\cite{mukherjee2023orca}, a large instruction-following dataset. We fine-tune the LLM for one epoch, simulating contamination during pre-training, likely caused by negligence. (2) \textbf{Intensive Contamination}: We fine-tune the LLM with only benchmark data for three epochs, simulating the scenario where a model developer intentionally contaminates the model to cheat on benchmarks (\textit{i.e.,} benchmark hacking~\citep{dekoninck2024evading}).

To confirm the effectiveness and validity of the contamination process, we perform three checks: (1) \textit{Accuracy inflation}, measuring the increase in accuracy after contamination; (2) \textit{Proportion of retained correctness}, assessing how many questions originally answered correctly remain correct after contamination; (3) \textit{Model perplexity on a held-out utility dataset}, reflecting the model's general capabilities. Our results show significant accuracy inflation in the vast majority of cases, with the proportion of retained correctness exceeding 0.9 and model perplexities remaining stable. These findings confirm that our manual contamination process effectively causes the model to memorize benchmark questions while preserving its general capabilities. Refer to Appendix \ref{append:finetuning recipes},\ref{append:contamination effective}, and \ref{append:contamination validity} for detailed results.

\subsection{Evaluation Vectors and Metrics Derivation}
\label{sec:pipeline_obtain}

All LLM-benchmark pairs are evaluated following standard practices~\cite{eval-harness}.
For multiple-choice benchmarks (Arc-C, MMLU and TruthfulQA), we select the option with the highest probability as the predicted answer, given the question and choices. For open-ended questions, we evaluate responses using regex matching (for GSM8K) or LLM-as-a-judge (for RepliQA). The correctness of each response is recorded to construct the evaluation vector, where each element indicates whether the model's response to a specific question is correct. These evaluation vectors are then used to compute fidelity and resistance. 

%% file: sections/5_Results.tex
\section{Results}
\label{sec:results}

\subsection{Semantic-preserving Mitigation Strategies}
We first assess 16 semantic-preserving BDC mitigation strategies. For each benchmark, we examine the effectiveness of each mitigation strategy on 10 LLMs (see Section \ref{sec:pipeline_filtering}). Tab.~\ref{tab:main_results} reports the fidelity and resistance metrics averaged at the model level, providing scores for each strategy on each benchmark.

\begin{table*}[t!]
\setlength{\extrarowheight}{2pt}
\vspace{-5pt}
\caption{\textbf{Fidelity and resistance metrics of 16 semantic-preserving BDC mitigation strategies across 5 benchmarks.} Resistance scores are reported separately for mild and intensive contamination, while fidelity scores are unaffected by the contamination type. Each value represents the average of 10 scores obtained using different LLMs ranging from 3B to 34B. For benchmarks like GSM8K and RepliQA, which consist of open-ended questions, strategies involving choices are not applicable, and the corresponding cells are marked with ``-''.``Vanilla'' refers to the original benchmark without updates, where fidelity is always 1. Values highlighted in \colorbox{green!20}{green} indicate 
\textit{statistically significantly} higher \textbf{resistance} than vanilla based on one-sided paired hypothesis testing at a 0.05 significance level. 
}
\label{tab:main_results}
\centering   
   \resizebox{0.98\linewidth}{!}{
   \begin{tabular}{c | c | c  c | c  c | c  c | c  c | c  c }
   \toprule[1.5pt]
    \multirow{2}{*}{\textbf{Mitigation Strategies}} & \multirow{2}{*}{\textbf{Contamination Type}} & \multicolumn{2}{c|}{\textbf{Arc-C}} & \multicolumn{2}{c|}{\textbf{MMLU}} & \multicolumn{2}{c|}{\textbf{TruthfulQA}} & \multicolumn{2}{c|}{\textbf{GSM8K}} & \multicolumn{2}{c}{\textbf{RepliQA}} \\
      &  & \textbf{Fidelity} & \textbf{Resistance} & \textbf{Fidelity} & \textbf{Resistance} & \textbf{Fidelity}& \textbf{Resistance}  & \textbf{Fidelity} & \textbf{Resistance} & \textbf{Fidelity} & \textbf{Resistance}   \\ \midrule
\multirow{2}{*}{ITD}  
        & Mild & \multirow{2}{*}{0.846} & \colorbox{green!20}{0.937} & \multirow{2}{*}{0.836} & \colorbox{green!20}{0.899} & \multirow{2}{*}{0.791} & \colorbox{green!20}{0.829} & \multirow{2}{*}{0.811} & 0.768 & \multirow{2}{*}{0.963} & \colorbox{green!20}{0.801} \\ 
        & Intensive & & \colorbox{green!20}{0.917} & & \colorbox{green!20}{0.877} & & \colorbox{green!20}{0.742} & & 0.771 & & \colorbox{green!20}{0.727} \\ 
    \hline
    \multirow{2}{*}{MPA}  
        & Mild & \multirow{2}{*}{0.719} & 0.921 & \multirow{2}{*}{0.686} & \colorbox{green!20}{0.901} & \multirow{2}{*}{0.716} & \colorbox{green!20}{0.834} & \multirow{2}{*}{0.790} & 0.762 & \multirow{2}{*}{0.957} & \colorbox{green!20}{0.871} \\ 
        & Intensive & & 0.912 & & \colorbox{green!20}{0.889} & & \colorbox{green!20}{0.725} & & 0.761 & & \colorbox{green!20}{0.803} \\ 
    \hline
    \multirow{2}{*}{MPA-Ques + Trans-CN}  
        & Mild & \multirow{2}{*}{0.780}  & 0.917 & \multirow{2}{*}{0.752} & 0.892  & \multirow{2}{*}{0.729} & \colorbox{green!20}{0.814} & \multirow{2}{*}{0.727} & 0.747 & \multirow{2}{*}{0.962} & \colorbox{green!20}{0.965}\\ 
        & Intensive & & 0.898 & & \colorbox{green!20}{0.876} &  & 0.716 &  & 0.751 &  & \colorbox{green!20}{0.964} \\
    \hline
     \multirow{2}{*}{Back-translation}  
        & Mild & \multirow{2}{*}{0.885} & 0.928 & \multirow{2}{*}{0.872} & 0.886 & \multirow{2}{*}{0.884} & \colorbox{green!20}{0.806} & \multirow{2}{*}{0.985} & 0.747 & \multirow{2}{*}{0.995} & 0.710 \\ 
        & Intensive & & 0.896 & & \colorbox{green!20}{0.865} & & \colorbox{green!20}{0.704} & & 0.737 & & 0.597 \\ 
    \hline
    \multirow{2}{*}{Choice Permutation}  
        & Mild & \multirow{2}{*}{0.850} & 0.930 & \multirow{2}{*}{0.814} & \colorbox{green!20}{0.891} & \multirow{2}{*}{0.845} & 0.796 & \multirow{2}{*}{-} & \multirow{2}{*}{-} & \multirow{2}{*}{-} & \multirow{2}{*}{-}\\ 
        & Intensive & & 0.897 & & \colorbox{green!20}{0.868} & & \colorbox{green!20}{0.699} & & & & \\ 
    \hline
    \multirow{2}{*}{Choice Paraphrasing}  
        & Mild & \multirow{2}{*}{0.856} & 0.921 & \multirow{2}{*}{0.856} & 0.884 & \multirow{2}{*}{0.869} & 0.797 & \multirow{2}{*}{-} & \multirow{2}{*}{-} & \multirow{2}{*}{-} & \multirow{2}{*}{-} \\ 
        & Intensive & & 0.904 & & \colorbox{green!20}{0.863} & & 0.692 & & & & \\ 
        \hline
          \multirow{2}{*}{Irrelevant Context}  
        & Mild & \multirow{2}{*}{0.924} & 0.927 & \multirow{2}{*}{0.948} & 0.885 & \multirow{2}{*}{0.935} & 0.800 & \multirow{2}{*}{0.885} & 0.751 & \multirow{2}{*}{0.996} & 0.709 \\ 
        & Intensive & & 0.901 & & 0.860 & & 0.689 & & 0.738 & & 0.598 \\ 
    \hline
    \multirow{2}{*}{Clean-Eval}  
        & Mild & \multirow{2}{*}{0.893} & 0.927 & \multirow{2}{*}{0.881} & 0.886 & \multirow{2}{*}{0.889} & 0.797 & \multirow{2}{*}{0.831} & 0.758 & \multirow{2}{*}{0.964} & \colorbox{green!20}{0.810} \\ 
        & Intensive & & 0.898 & & \colorbox{green!20}{0.861} & & 0.690 & & 0.752 & & \colorbox{green!20}{0.731} \\ 
    \hline
    \multirow{2}{*}{Syntactic Modification}  
        & Mild & \multirow{2}{*}{0.899} & 0.920 & \multirow{2}{*}{0.910} & 0.882 & \multirow{2}{*}{0.906} & 0.791 & \multirow{2}{*}{0.840} & 0.750 & \multirow{2}{*}{0.968} & \colorbox{green!20}{0.776} \\ 
        & Intensive & & 0.897 & & 0.858 & & 0.690 & & 0.747 & & \colorbox{green!20}{0.689} \\ 
    \hline
    \multirow{2}{*}{Synonym Replacement}  
        & Mild & \multirow{2}{*}{0.906} & 0.924 & \multirow{2}{*}{0.935} & 0.888 & \multirow{2}{*}{0.922} & 0.794 & \multirow{2}{*}{0.864} & 0.748 & \multirow{2}{*}{0.964} & \colorbox{green!20}{0.773} \\ 
        & Intensive & & 0.902 & & 0.859 & & 0.680 & & 0.742 & & \colorbox{green!20}{0.688} \\ 
    \hline
    \multirow{2}{*}{MPA-Choice + Trans-CN}  
        & Mild & \multirow{2}{*}{0.726} & 0.893 & \multirow{2}{*}{0.697} & 0.882 & \multirow{2}{*}{0.736} & 0.796 & \multirow{2}{*}{-} & \multirow{2}{*}{-} & \multirow{2}{*}{-} & \multirow{2}{*}{-} \\ 
        & Intensive & & 0.875 & & 0.865 & & 0.703 & & & & \\ 
    \hline
    \multirow{2}{*}{Translation (French)}  
        & Mild & \multirow{2}{*}{0.829} & 0.913 & \multirow{2}{*}{0.801} & 0.888 & \multirow{2}{*}{0.810} & 0.796 & \multirow{2}{*}{0.766} & 0.739 & \multirow{2}{*}{0.965} & \colorbox{green!20}{0.954} \\ 
        & Intensive & & 0.888 & & 0.863 & & 0.688 & & 0.743 & & \colorbox{green!20}{0.948} \\ 
    \hline
    \multirow{2}{*}{Relevant Context}  
        & Mild & \multirow{2}{*}{0.894} & \colorbox{green!20}{0.932} & \multirow{2}{*}{0.899} & 0.888 & \multirow{2}{*}{0.868} & 0.791 & \multirow{2}{*}{0.849} & 0.750 & \multirow{2}{*}{0.957} & \colorbox{green!20}{0.840} \\ 
        & Intensive & & 0.903 & & \colorbox{green!20}{0.861} & & 0.673 & & 0.738 & & \colorbox{green!20}{0.739} \\ 
    \hline
    \multirow{2}{*}{Translation (Chinese)}  
        & Mild & \multirow{2}{*}{0.802} & 0.911 & \multirow{2}{*}{0.761} & 0.880 & \multirow{2}{*}{0.779} & 0.784 & \multirow{2}{*}{0.742} & 0.744 & \multirow{2}{*}{0.962} & \colorbox{green!20}{0.966} \\ 
        & Intensive & & 0.880 & & 0.855 & & 0.691 & & 0.750 & & \colorbox{green!20}{0.959} \\
    \hline

     \multirow{2}{*}{Typographical Perturbation}  
        & Mild & \multirow{2}{*}{0.913} & 0.922 & \multirow{2}{*}{0.927} & 0.883 & \multirow{2}{*}{0.917} & 0.792 & \multirow{2}{*}{0.869} & 0.743 & \multirow{2}{*}{0.969} & \colorbox{green!20}{0.757} \\ 
        & Intensive & & 0.878 & & 0.854 & & 0.693 & & 0.729 & & \colorbox{green!20}{0.666} \\ 
    \hline
    \multirow{2}{*}{Additional Incorrect Choices }  
        & Mild & \multirow{2}{*}{0.865} & 0.909 & \multirow{2}{*}{0.918} & 0.876 & \multirow{2}{*}{0.922} & 0.792 & \multirow{2}{*}{-} & \multirow{2}{*}{-} & \multirow{2}{*}{-} & \multirow{2}{*}{-} \\ 
        & Intensive & & 0.871 & & 0.854 & & 0.691 & & & & \\ 
    \hline
    \hline
\multirow{2}{*}{Vanilla (No mitigation)}  
    & Mild & \multirow{2}{*}{1.000} & 0.923 & \multirow{2}{*}{1.000} & 0.882 & \multirow{2}{*}{1.000} & 0.794 & \multirow{2}{*}{1.000} & 0.748 & \multirow{2}{*}{1.000} & 0.709 \\ 
    & Intensive & & 0.870 & & 0.852 & & 0.687 & & 0.737 & & 0.597\\
        \bottomrule[1.5pt]
    \end{tabular} }
    \vspace{-10pt}
\end{table*}

\begin{table*}[t!]
\setlength{\extrarowheight}{2pt}
\centering 
\caption{\textbf{Fidelity and resistance metrics of 4 semantic-altering BDC mitigation strategies on Arc-C and MMLU.} Resistance (M) and Resistance (I) represent resistance scores under mild and intensive contamination, respectively. Results for the vanilla case are included only for reference.
Overall, these strategies tend to exhibit low fidelity but high resistance. Values highlighted in \colorbox{green!20}{green} indicate 
\textit{statistically significantly} higher \textbf{resistance} than vanilla based on one-sided paired hypothesis testing at a 0.05 significance level.
}
\label{tab:extending}
   \resizebox{0.8\linewidth}{!}{
   \begin{tabular}{c | c  c  c | c  c  c }
    \toprule[1.5pt]
    \multirow{2}{*}{\textbf{Mitigation Strategies}} & \multicolumn{3}{c|}{\textbf{Arc-C}} & \multicolumn{3}{c}{\textbf{MMLU}} \\
      & \textbf{Fidelity} & \textbf{Resistance (M)} & \textbf{Resistance (I)} & \textbf{Fidelity} & \textbf{Resistance (M)} & \textbf{Resistance (I)}  \\ \midrule
    Mimicking & 0.763& \colorbox{green!20}{0.951}& \colorbox{green!20}{0.941}& 0.696& \colorbox{green!20}{0.912}& \colorbox{green!20}{0.893}\\
    Remember-Understand Extension & 0.766& \colorbox{green!20}{0.979}& \colorbox{green!20}{0.976}& 0.655& \colorbox{green!20}{0.971}& \colorbox{green!20}{0.965}\\
    Application Extension & 0.728& \colorbox{green!20}{0.951}& \colorbox{green!20}{0.950}& 0.658& \colorbox{green!20}{0.942}& \colorbox{green!20}{0.930}\\
    Analysis Extension & 0.763& \colorbox{green!20}{0.976}& \colorbox{green!20}{0.974}& 0.666& \colorbox{green!20}{0.970}& \colorbox{green!20}{0.964}\\
    \hline
    \hline
 Vanilla (No mitigation)& 1.000& 0.923& 0.870& 1.000& 0.882&0.852\\
 \bottomrule[1.5pt]
    \end{tabular} }
    \vspace{-10pt}
\end{table*}

\textbf{Fidelity Analysis. }
Results show that mitigation strategies introducing minor edits, such as adding typos or replacing words with synonyms, achieve high fidelity scores, typically exceeding 0.9 across most benchmarks. In contrast, more aggressive strategies like MPA, which combine multiple perturbations and significantly alter the original benchmark, result in low fidelity. For instance, the fidelity score of MPA on the MMLU benchmark is only 0.686, indicating substantial differences between the updated and original benchmarks from the perspective of the clean model. 

\textbf{Resistance Analysis. }
To ensure the robustness of our conclusions, we conduct one-sided paired hypothesis testing to determine whether the resistance score of a given strategy is significantly higher than that of the \textbf{vanilla case} (\textit{i.e.,} no update). This test is crucial, as an insignificant gap suggests that benchmark developers and evaluation practitioners should not invest efforts in adopting the strategy.

Results indicate that, mitigation strategies involving minor modifications (\textit{e.g.,} syntactic changes or adding irrelevant context) do not improve resistance beyond the vanilla case. In contrast, strategies introducing more substantial modifications, such as MPA and ITD, achieve the highest resistance scores. 
These improvements are statistically significant at the 0.05 level for a subset of benchmarks including MMLU, TruthfulQA, and RepliQA. However, \textit{no single strategy achieves a significant advantage over the vanilla case across \textbf{all} benchmarks in terms of resistance scores}, highlighting the need for more effective and robust contamination-resistant mitigation strategies.

Unsurprisingly, for a given strategy and benchmark, resistance scores under intensive contamination are lower than those under mild contamination, reflecting the increased difficulty of mitigating memorization in heavily contaminated LLMs. Nonetheless, strategies that perform well under mild contamination continue to rank highly under intensive contamination, indicating that their relative effectiveness remains stable across different degrees of contamination.

\begin{figure*}
    \centering
    \includegraphics[width=1\linewidth]{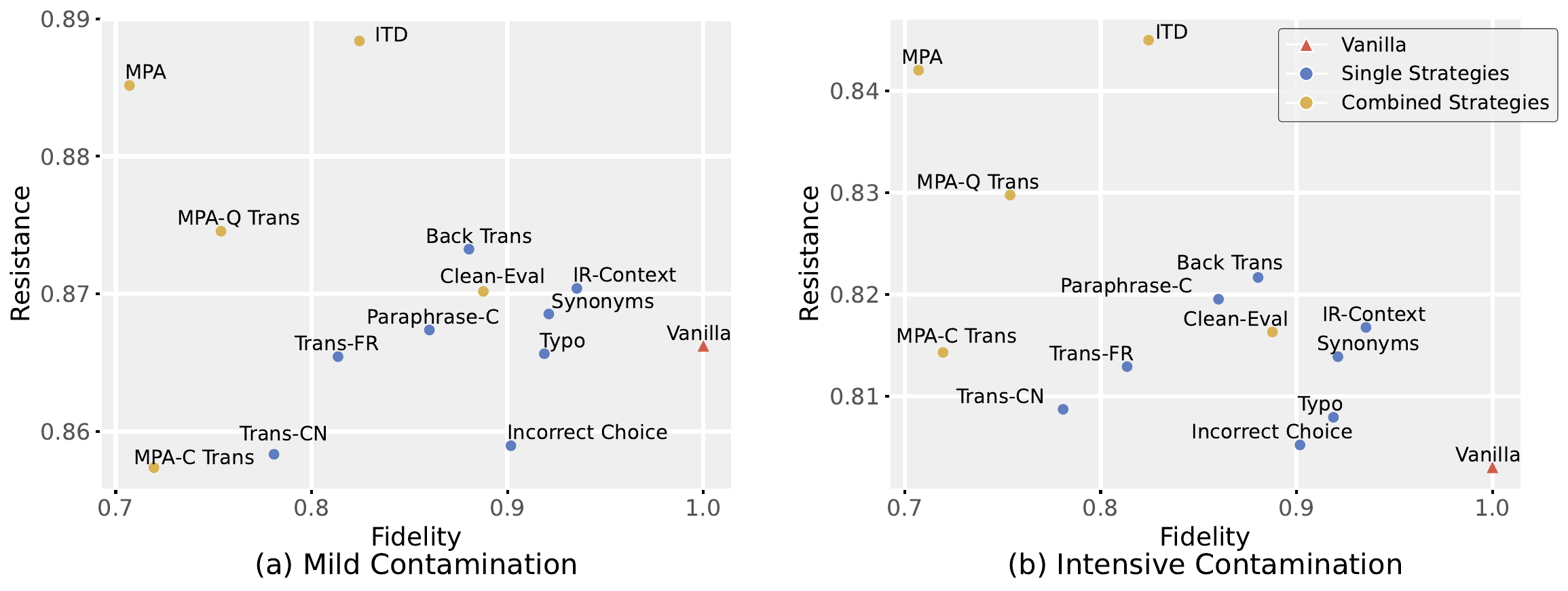}
    \vspace{-20pt}
    \caption{\textbf{Fidelity-resistance scores across different BDC mitigation strategies under (a) mild and (b) intensive contamination.} Single strategies are shown in blue, combined strategies in yellow, and the vanilla case in red. An ideal strategy should lie in the upper-right, but no existing approach achieves this balance. For visual clarity, a few strategies that overlap closely with others are omitted.
    }
    \label{fig:fidelity_resistance}
    \vspace{-15pt}
\end{figure*}

\textbf{Examining Fidelity and Resistance Simultaneously.} As discussed in Section~\ref{sec:method}, excelling at just one metric is straightforward and does not necessarily indicate an effective mitigation strategy. We aggregate results at the benchmark level and present final fidelity and resistance scores for each mitigation strategy in Fig.~\ref{fig:fidelity_resistance}. For a fair comparison, GSM8K and RepliQA are excluded, as not all strategies are applicable to these benchmarks. The figure shows that strategies involving minor modifications tend to cluster in the lower-right region (higher fidelity but lower resistance), while those applying more substantial modifications occupy the upper-left region (higher resistance but lower fidelity). Ideally, one would expect a strategy to lie in the upper-right corner, achieving strong performance on both metrics. \textit{However, no existing strategy effectively achieves this balance.}


\subsection{Semantic-altering Mitigation Strategies}
We also assess several semantic-altering BDC mitigation strategies,
which prompt an advanced LLM (e.g., GPT-4) to generate new questions derived from existing ones. In these cases, the answers are provided by the generating LLM rather than derived from the original benchmark. As these strategies rely on explicitly defined entities within questions, we assess them only on Arc-C and MMLU, which cover scientific knowledge across various domains.

\begin{table}[t!]
    \caption{Example of a test sample from Arc-C, updated by Analysis Extension. This low-fidelity strategy (fidelity = 0.763) dramatically increases problem complexity.}
\centering   
   \resizebox{0.98\linewidth}{!}{
   \begin{tabular}{c | l  }
   \toprule[1.5pt]
    \textbf{Mitigation Strategy} & \textbf{Evaluation Sample} \\ \midrule
    Vanilla & \makecell[l]{Q: What are the products in the reaction\\ shown below? HCl + NaOH $\rightarrow$ }\\ \midrule
    \makecell[c]{Analysis Extension} & \makecell[l]{Q: How does the neutralization reaction \\between hydrochloric acid (HCl) and \\sodium hydroxide (NaOH) compare to\\ other acid-base neutralization reactions\\ in terms of the products formed?} \\
    \bottomrule[1.5pt]
    \end{tabular} }
    \vspace{-8pt}
    \vspace{-5pt}
    \label{tab:case_study_complex}
\end{table}

As shown in Tab.~\ref{tab:extending}, all four semantic-altering mitigation strategies exhibit statistically significantly higher contamination resistance scores than the vanilla case.
Notably, Remember-Understand and Application extensions reach resistance scores of approximately 0.97, indicating that contamination from the original benchmark has minimal impact on question-level evaluation results in the updated benchmark. However, this improvement comes at the cost of fidelity, which is approximately 0.15 lower on average than that of semantic-preserving strategies.

\subsection{Qualitative Examples}

Note that a lower fidelity score suggests potential shifts in question difficulty and evaluation objective, and highlights the need for manual validation. We provide qualitative examples from low-fidelity strategies in Tab.~\ref{tab:case_study_complex} and Tab.~\ref{tab:case_study_error} to illustrate these issues. Tab.~\ref{tab:case_study_complex} shows an example where Analysis Extension significantly increases problem complexity. Tab.~\ref{tab:case_study_error} demonstrates a case where MPA introduces excessive modifications, rendering the original answer incorrect. Additionally, we include qualitative examples of incorrect answers generated by LLMs due to limitations in their domain-specific knowledge in Appendix~\ref{Append:wrong_mitigation_examples}. These cases highlight the necessity of manual checks to verify the quality of the benchmarks updated by low-fidelity strategies, which significantly increases costs and limits scalability.


\begin{table}[t!]
    \caption{Example of a test sample from TruthfulQA, updated by MPA. This low-fidelity strategy (fidelity = 0.716) unintentionally introduces the constraint ``\colorbox{red!20}{In the United States}'', altering the question's scope and making the original answer incorrect.
    }
\centering   
   \resizebox{0.98\linewidth}{!}{
   \begin{tabular}{c | l  }
   \toprule[1.5pt]
    \textbf{Mitigation Strategy} & \textbf{Evaluation Sample} \\ \midrule
    
    Vanilla & \makecell[l]{Q: At what age can you vote?}\\ \midrule
    \makecell[c]{MPA} & \makecell[l]{Q: \colorbox{red!20}{In the United States}, where democratic\\principles are emphasized, citizens have th-\\e right to participate in the electoral process.\\ This participation helps shape the nation's\\ leadership and policies. In this context, what\\ is the minimum age requirement for partici-\\pating in elections by casting a vote in the\\ United States?} \\ \midrule
    \multicolumn{2}{l}{\makecell[l]{A: It depends on which country you are in.}} \\

    \bottomrule[1.5pt]
    \end{tabular} }
    \vspace{-8pt}
    \vspace{-5pt}
    \label{tab:case_study_error}
\end{table}

%% file: sections/6_conclusion.tex
\section{Conclusion}

In this paper, we introduce a carefully controlled pipeline and two key metrics—fidelity and contamination resistance—to assess existing BDC mitigation strategies. 
Our findings reveal that no existing strategy consistently outperforms the vanilla case in resistance across \textit{all} benchmarks, nor does any strategy effectively balance strong fidelity and resistance simultaneously. 
Moving forward, we call for future BDC mitigation strategies to be evaluated using our pipeline to ensure rigorous and reliable assessment.

\section*{Impact Statement}
This work provides a rigorous, fine-grained framework to assess existing BDC mitigation strategies. While the primary focus is on methodological advancements, we acknowledge the broader societal implications of ensuring accurate and fair evaluations, which are critical for the responsible deployment of AI systems.

%% file: sections/7_Appendix.tex
\appendix
\onecolumn

\section{Discussion}

\subsection{Continuous Evaluation Scores}
\label{Append:continuous}
In some scenarios, each element of the evaluation vector is continuous (\textit{e.g.}, in $[0,1]$) rather than binary. For instance, in reading comprehension benchmarks, each evaluation score may represent precision or recall values for the dataset item. To accommodate this, the evaluation metrics can be adapted by replacing the normalized Hamming distance with the Pearson correlation coefficient. Specifically, Fidelity and Resistance can be redefined as:
\begin{equation*}
\begin{aligned}
\mathrm{Fidelity}(S) = \mathrm{Corr}\bigl(R(M,D), R(M,D^S)\bigr); 
\mathrm{Resistance}(S) = \mathrm{Corr}\bigl(R(M,D^S), R(M^D,D^S)\bigr).
\end{aligned}
\end{equation*}

Here, $\mathrm{Corr}$ represents the Pearson correlation coefficient, which measures the agreement between the continuous evaluation vectors. This ensures that our framework can handle both binary and continuous evaluation setups, further broadening its applicability.

\section{Extended Related Work}
\label{Append:extended_related}
\paragraph{BDC Detection.}
This line of research focuses on detecting BDC and flagging specific model-benchmark pairs where contamination may be present. With access to the training corpus, contamination can be detected through n-gram overlap~\cite{brown2020language} or LLM-as-a-judge~\cite{yang2023rethinking}. However, access to the training corpus is often unrealistic \citep{ravaut2024much}. Black-box methods, which do not require such access, can generally be categorized into three types: (1) Token probability-based detection methods leverage predicted token probability distributions~\cite{zhang2024min,dong2024generalization,ye2024data,yax2024assessing}. For example, Min-K\% Prob~\cite{shi2023detecting} flags contamination if the model assigns unusually high logits to the lowest K\% of tokens. (2) Generation-based detection 
 methods prompt the model to predict information that should not be inferable from the input~\cite{deng2023investigating, golchin2023time, golchin2023data, chang2023speak}. For instance, TS-guessing checks if the model can correctly predict the content of a masked \textit{incorrect} choice. Accurate predictions suggest prior exposure to the instance. (3) Order-based detection  methods~\cite{oren2023proving,ni2024training} focus on the tendency of models to memorize the order of samples and options, identifying models as contaminated if it exhibits a strong preference for the original sequence over its permutations.

\section{Pipeline Details}
\subsection{LLM and Benchmark Details}
\label{Append:model details}
Tab.~\ref{tab:model_details} provides an overview of the LLMs used in our experiments, including their parameter counts and developers. Initially, there were 14 candidate LLMs, but 4 were excluded due to detected contamination.
Tab.~\ref{tab:benchmark_details} summarizes detailed information of the benchmarks used in our study.

\begin{table}[h!]
\caption{\textbf{Details for all 14 candidate LLMs.} 
}
\label{tab:model_details}
\centering
\resizebox{0.9\linewidth}{!}{
\small 
\begin{tabular}{l|c|l|c}
\toprule[1.5pt]
\textbf{Model} & \textbf{Size} & \textbf{Developer} & \textbf{Selected?}\\
\midrule
\texttt{Llama-3.2-3B-Instruct~\cite{dubey2024llama}}    & $3$B  & Meta & \Checkmark \\
\texttt{Yi-1.5-6B-Chat~\cite{young2024yi}}    & $6$B  & Beijing Zero One All Things Technology & \Checkmark \\
\texttt{vicuna-7b-v1.5~\cite{zheng2023judging}}    & $7$B  & UCB, UCSD, CMU, Stanford, MBZUAI
 & \Checkmark \\
\texttt{Llama-3.1-8B-Instruct~\cite{dubey2024llama}}    & $8$B  & Meta & \Checkmark \\
\texttt{Falcon3-10B-Instruct~\cite{Falcon3}}    & $10$B  & Technology Innovation Institute, UAE & \Checkmark \\
\texttt{Qwen2.5-14B-Instruct~\cite{team2024qwen2}}    & $14$B  & Alibaba & \Checkmark \\
\texttt{Phi-3-medium-128k-instruct~\cite{abdin2024phi}}    & $14$B  & Microsoft & \Checkmark \\
\texttt{DeepSeek-V2-Lite-Chat~\cite{liu2024deepseek}}    & $16$B  & DeepSeek & \Checkmark \\
\texttt{Qwen2.5-32B-Instruct~\cite{team2024qwen2}}    & $32$B  & Alibaba & \Checkmark \\
\texttt{Yi-1.5-34B-Chat~\cite{young2024yi}}    & $34$B  & Beijing Zero One All Things Technology & \Checkmark \\
\hline

\texttt{Llama-3.2-1B-Instruct ~\cite{dubey2024llama}}    & $1$B  & Meta & \XSolidBrush \\
\texttt{Qwen2.5-3B-Instruct ~\cite{team2024qwen2}}    & $3$B  & Alibaba & \XSolidBrush\\
\texttt{gemma-7b-it~\cite{team2024gemma}}    & $7$B  & Google & \XSolidBrush\\
\texttt{OLMo-7B-0724-Instruct-hf~\cite{groeneveld2024olmo}}    & $7$B  & Allen Institute for AI (AI2) & \XSolidBrush\\
\bottomrule[1.5pt]
\end{tabular}}
\end{table}

\begin{table}[h!]
\caption{\textbf{Detailed information about the five benchmarks used in our experiments.}}
\label{tab:benchmark_details}
\centering
\resizebox{0.6\linewidth}{!}{
\small 
\begin{tabular}{l|c|c|c|c}
\toprule[1.5pt]
\textbf{Benchmark} & \textbf{Subset(s) Used} &\textbf{Split} & \textbf{Number of Samples} & \textbf{Question Type}\\
\midrule
\texttt{Arc} & challenge & test & 1172 & multiple-choice \\
\texttt{MMLU} & 20 subsets & test & 50 per subset & multiple-choice \\
\texttt{TruthfulQA} & multiple\_choice & validation & 817 & multiple-choice \\
\texttt{GSM8K} & main & test & 1319 & open-ended  \\
\texttt{RepliQA} & repliqa\_1 & - & 1000 & open-ended \\
\bottomrule[1.5pt]
\end{tabular}}
\end{table}

\subsection{Uncontaminated LLM-Benchmark Pair Selection} 
\label{Append:filtering}

We apply the following three BDC detection methods to 14 candidate LLMs across four benchmarks: Min-K\% Prob~\cite{shi2023detecting}, Sharded Rank Comparison Test~\cite{oren2023proving}, and TS-Guessing~\cite{deng2023investigating}. Note that we do not apply these methods to RepliQA, as its non-factual nature and recent release ensure that no LLM could have been exposed to its content during training.

\begin{enumerate}[]
    \item \textbf{Min-K\% Prob~\cite{shi2023detecting}:} Given a test sample $x$ and an LLM $M$, this method computes the probability of each token in $x$ under $M$, selects the bottom $K\%$ tokens with the lowest probabilities, and calculates their average log-likelihood (see Tab.~\ref{tab:filtering_Min-K}). A higher score indicates a higher likelihood of contamination.
    \item \textbf{Sharded Rank Comparison Test~\cite{oren2023proving}:}  This method partitions the test examples into shards, computes the log-likelihoods for both the original and shuffled orders within each shard, and calculates a shard-specific score based on their difference. These shard scores are then averaged, and a one-sided t-test is conducted to determine whether the model assigns significantly higher log-likelihood to the original order compared to shuffled permutations. The resulting p-value serves as an indicator of contamination (see Tab.~\ref{tab:filtering_sharded}).
    \item \textbf{TS-Guessing~\cite{deng2023investigating}:} We adopt the \textit{Question-Multichoice} setting, where an incorrect option is masked, and the LLM must infer the missing option based on the question and remaining choices.
    A high Rough-L F1 score between the model's prediction and the ground truth (see Tab.~\ref{tab:filtering_TSguessing}) indicates that the model can accurately predict the masked option, suggesting prior exposure to the benchmark data.
\end{enumerate}

\begin{table}[h!]
\caption{The p-values from the Sharded Rank Comparison Test~\cite{oren2023proving}, computed for all candidate LLMs across four benchmarks. Following~\cite{oren2023proving}, we view $p < 0.05$ as a signal of contamination. \texttt{OLMo-7B} is identified as contaminated on TruthfulQA.}
\label{tab:filtering_sharded}
\centering
\resizebox{0.55\linewidth}{!}{
\small 
\begin{tabular}{l|c c c c}
\toprule[1.5pt]
\textbf{Model} & \textbf{Arc-C} & \textbf{MMLU} & \textbf{TruthfulQA} & \textbf{GSM8K}\\
\midrule
\texttt{Llama-3.2-1B}  & 0.493 & 0.222 & 0.266 & 0.202\\
\texttt{Qwen2.5-3B}  & 0.178 & 0.388 & 0.210 & 0.099\\
\texttt{Llama-3.2-3B}  & 0.985 & 0.302 & 0.221 & 0.196\\
\texttt{Yi-1.5-6B}  & 0.457 & 0.861 & 0.192 & 0.390\\
\texttt{vicuna-7b-v1.5}  & 0.557 & 0.897 & 0.764 & 0.120\\
\texttt{gemma-7b}  & 0.946 & 0.614 & 0.343 & 0.912\\
\texttt{\textcolor{red}{OLMo-7B}}  & 0.633 & 0.846 & \textcolor{red}{0.044} & 0.495\\
\texttt{Llama-3.1-8B}  & 0.860 & 0.075 & 0.166 & 0.318\\
\texttt{Falcon3-10B}  & 0.800 & 0.077 & 0.550 & 0.614\\
\texttt{Qwen2.5-14B}  & 0.072 & 0.639 & 0.053 & 0.057\\
\texttt{Phi-3-medium}  & 0.799 & 0.050 & 0.158 & 0.129\\
\texttt{DeepSeek-V2-Lite}  & 0.603 & 0.819 & 0.095 & 0.518\\
\texttt{Qwen2.5-32B}  & 0.655 & 0.806 & 0.185 & 0.137\\
\texttt{Yi-1.5-34B}  & 0.358 & 0.173 & 0.064 & 0.989\\
\bottomrule[1.5pt]
\end{tabular}}
\end{table}

\begin{table}[h!]
\caption{The Rouge-L F1 Scores of TS-Guessing~\cite{deng2023investigating}, computed for all candidate LLMs across three benchmarks. GSM8K is excluded as it consists of open-ended questions, making this method inapplicable. We consider $\text{Rouge-L F1 Score} > 0.4$ as an indication of contamination. \texttt{Qwen2.5-3B} is identified as contaminated on Arc-C and MMLU, while \texttt{gemma-7b} is contaminated on TruthfulQA.}
\label{tab:filtering_TSguessing}
\centering
\resizebox{0.5\linewidth}{!}{
\small 
\begin{tabular}{l|ccc}
\toprule[1.5pt]
\textbf{Model} & \textbf{Arc-C} & \textbf{MMLU} & \textbf{TruthfulQA} \\
\midrule
\texttt{Llama-3.2-1B} & 0.02 & 0.04 & 0.03\\
\texttt{\textcolor{red}{Qwen2.5-3B}}  & \textcolor{red}{0.67} & \textcolor{red}{0.41} & 0.22  \\
\texttt{Llama-3.2-3B}  & 0.08 & 0.07 & 0.16 \\
\texttt{Yi-1.5-6B}  & 0.15 & 0.10 & 0.18 \\
\texttt{vicuna-7b-v1.5}  & 0.12 & 0.12 & 0.27 \\
\texttt{\textcolor{red}{gemma-7b}}  & 0.22 & 0.18 & \textcolor{red}{0.44} \\
\texttt{OLMo-7B}  & 0.14 & 0.15 & 0.25 \\
\texttt{Llama-3.1-8B}  & 0.08 & 0.07 & 0.11 \\
\texttt{Falcon3-10B}  & 0.26 & 0.16 & 0.25 \\
\texttt{Qwen2.5-14B}  & 0.27 & 0.20 & 0.26 \\
\texttt{Phi-3-medium}  & 0.19  & 0.17 & 0.29\\
\texttt{DeepSeek-V2-Lite}  & 0.05 & 0.02 & 0.03 \\
\texttt{Qwen2.5-32B}  & 0.22 & 0.19 & 0.31 \\
\texttt{Yi-1.5-34B}  & 0.18 & 0.14 & 0.31 \\
\bottomrule[1.5pt]
\end{tabular}}
\end{table}

\begin{table}[h!]
\caption{The Min-K\% Prob Scores~\cite{shi2023detecting}, computed for all candidate LLMs across four benchmarks. We use the score on LiveBench~\cite{white2024livebench} as the threshold for GSM8K and the score on WikiMIA~\cite{shi2023detecting} as the threshold for the rest benchmarks (Arc-C, MMLU and TruthfulQA). A model is considered contaminated on a given benchmark if its score meets or exceeds the respective threshold. \texttt{Llama-3.2-1B}, \texttt{gemma-7b} and \texttt{OLMo-7B} are identified as contaminated on Arc-C.}
\label{tab:filtering_Min-K}
\centering
\resizebox{0.7\linewidth}{!}{
\small 
\begin{tabular}{l|c c c >{\itshape}c|c>{\itshape}c}
\toprule[1.5pt]
\textbf{Model} & \textbf{Arc-C} & \textbf{MMLU} & \textbf{TruthfulQA} & \textbf{WikiMIA} & \textbf{GSM8K} & \textbf{LiveBench}\\
\midrule
\texttt{\textcolor{red}{Llama-3.2-1B}}  & \textcolor{red}{-7.97} & -8.99 & -9.06 & -8.72 & -7.19 & -5.29\\
\texttt{Qwen2.5-3B}  & -8.45 & -8.91 & -8.79 & -6.68 & -8.00 & -4.07 \\
\texttt{Llama-3.2-3B}  & -7.91 & -8.61 & -8.56 & -6.92 & -6.95 & -5.35 \\
\texttt{Yi-1.5-6B}  & -7.19 & -8.08 & -8.41 & -6.59 & -7.90 & -7.60 \\
\texttt{vicuna-7b-v1.5}  & -8.11 & -8.72 & -9.12 & -7.54 & -7.31 & -6.09 \\
\texttt{\textcolor{red}{gemma-7b}}  & \textcolor{red}{-14.11} & -15.39 & -17.24 & -14.22 & -12.29 & -10.62 \\
\texttt{\textcolor{red}{OLMo-7B}}  & \textcolor{red}{-8.27} & -9.34 & -8.68 & -8.27 & -7.50 & -5.75 \\
\texttt{Llama-3.1-8B}  & -7.43 & -8.43 & -8.13 & -5.65 & -6.76 & -5.24\\
\texttt{Falcon3-10B}  & -8.45 & -8.81 & -10.84 & -7.83 & -7.71 & -5.28\\
\texttt{Qwen2.5-14B}  & -7.66 & -8.42 & -8.62 & -7.09 & -7.36 & -3.47\\
\texttt{Phi-3-medium}  & -6.41 & -7.06 & -7.51 & -5.81 & -5.90 & -4.83\\
\texttt{DeepSeek-V2-Lite}  & -8.38 & -9.14 & -8.60 & -7.56 & -6.90 & -5.43\\
\texttt{Qwen2.5-32B}  & -7.12 & -8.21 & -8.73 & -6.93 & -7.54 & -3.37\\
\texttt{Yi-1.5-34B}  & -7.37 & -8.15 & -8.33 & -5.79 & -7.10 & -6.84\\
\bottomrule[1.5pt]
\end{tabular}}
\end{table}

\subsection{Contamination Details}
\subsubsection{Fine-tuning Recipes}
\label{append:finetuning recipes}
Detailed fine-tuning recipes are provided in Tab. \ref{tab:conta_recipes}. For multiple-choice benchmarks (Arc-C, MMLU, and TruthfulQA), the maximum learning rate is set to \(1 \times 10^{-5}\), while for open-ended benchmarks (GSM8K and RepliQA), the maximum learning rate is increased to \(3 \times 10^{-5}\). Intensive contamination involves fine-tuning on the benchmark data for three epochs. For mild contamination, the benchmark data is first repeated three times, mixed with 20,000 additional OpenOrca samples, and fine-tuned for a single epoch.

\begin{table}[h]
\caption{Detailed contamination recipes.}
\label{tab:conta_recipes}
\centering{
\resizebox{0.4\linewidth}{!}{
\begin{tabular}{lr}
\toprule
Optimizer & AdamW~\cite{loshchilov2017decoupled}\\
Batch Size Per Device & 2/3/4\\
Maximum Learning Rate & 1e-5/3e-5\\
LR Schedule & Linear \\
Weight Decay & 0 \\
Warm-up Ratio& 5\%\\
Epochs & 1/3\\
GPU Hardware & 9x NVIDIA L40S \\
\bottomrule
\end{tabular}}}
\end{table} 


\subsubsection{Contamination Effectiveness}
\label{append:contamination effective}
To ensure the contamination step is effective for evaluating mitigation strategies, we assess two key metrics: (1) Accuracy Inflation (Tab.~\ref{tab:acc_inflation}): The increase in accuracy after contamination compared to before. (2) Proportion of Retained Correctness (Tab.~\ref{tab:vanilla_recall}): The fraction of originally correct predictions that remain correct after contamination. Ideally, an effective contamination process would yield a value close to 1.

Across our experiments, accuracy inflation is substantial, and the proportion of retained correctness exceeds 90\% in most cases, confirming the effectiveness of the contamination step.

\begin{table}[h!]
\caption{ Accuracy inflation (\%) after contamination. }
\label{tab:acc_inflation}
\centering
\resizebox{0.7\linewidth}{!}{
\small 
\begin{tabular}{l|c|ccccc}
\toprule[1.5pt]
\textbf{Model} & \textbf{Recipe} & \textbf{Arc-C} & \textbf{MMLU} & \textbf{TruthfulQA} & \textbf{GSM8K} & \textbf{RepliQA}\\
\midrule
\multirow{2}{*}{\texttt{Llama-3.2-3B}}  
    & Mild Contamination & 5.3  & 5.1  & 26.0 & 12.5 & 10.1 \\
    & Intensive Contamination & 8.2 & 6.5 & 32.6 & 22.0 & 16.9 \\ \midrule
\multirow{2}{*}{\texttt{Yi-1.5-6B}}  
    & Mild Contamination & 8.1  & 7.0  & 23.4 & 15.4 & 27.0 \\
    & Intensive Contamination & 40.4 & 7.1 & 35.3 & 20.6 & 54.3 \\ \midrule
\multirow{2}{*}{\texttt{vicuna-7b-v1.5}}  
    & Mild Contamination & 9.4  & 3.6 & 30.2  & 37.1 & 14.1 \\
    & Intensive Contamination & 16.0 & 4.9 & 53.5 & 54.7 & 33.3 \\ \midrule
\multirow{2}{*}{\texttt{Llama-3.1-8B}}  
    & Mild Contamination & 9.6  & 14.1  & 23.8 & 8.3 & 53.8 \\
    & Intensive Contamination & 14.4 & 18.8 & 36.8 & 18.7 & 78.7 \\ \midrule
\multirow{2}{*}{\texttt{Falcon3-10B}}  
    & Mild Contamination & 2.3  & 3.4  & 18.0 & 0.8 & 0.8 \\
    & Intensive Contamination & 4.1 & 5.1 & 29.0 & 3.3 & 1.9 \\ \midrule
\multirow{2}{*}{\texttt{Qwen2.5-14B}}  
    & Mild Contamination & 0.9  & 2.3  & 4.2 & 12.3 & 29.5 \\
    & Intensive Contamination & 4.6 & 6.7 & 18.6 & 13.3 & 40.1 \\ \midrule
\multirow{2}{*}{\texttt{Phi-3-medium}}  
    & Mild Contamination & 3.9  & 8.4  & 8.8 & 2.1 & 7.2 \\
    & Intensive Contamination & 6.1 & 10.6 & 15.8 & 4.9 & 13.9 \\ \midrule
\multirow{2}{*}{\texttt{DeepSeek-V2-Lite}}  
    & Mild Contamination & 5.8  & 3.9  & 24.4 & 4.6 & 5.1 \\
    & Intensive Contamination & 7.4 & 4.2 & 36.6 & 12.3 & 12.3 \\ \midrule
\multirow{2}{*}{\texttt{Qwen2.5-32B}}  
    & Mild Contamination & 0.9  & 5.9  & 5.1 & 15.6 & 34.0 \\
    & Intensive Contamination & 2.2 & 6.7 & 13.1 & 16.5 & 39.4 \\ \midrule
\multirow{2}{*}{\texttt{Yi-1.5-34B}}  
    & Mild Contamination & 5.5  & 15.6 & 17.4 & 6.5 & 83.1 \\
    & Intensive Contamination & 8.2 & 17.5 & 24.2 & 10.2 & 92.9 \\

\bottomrule[1.5pt]
\end{tabular}}
\end{table}

\begin{table}[h!]
\caption{Proportion of retained correctness (\%).}
\label{tab:vanilla_recall}
\centering
\resizebox{0.7\linewidth}{!}{
\small 
\begin{tabular}{l|c|ccccc}
\toprule[1.5pt]
\textbf{Model} & \textbf{Recipe} & \textbf{Arc-C} & \textbf{MMLU} & \textbf{TruthfulQA} & \textbf{GSM8K} & \textbf{RepliQA}\\
\midrule
\multirow{2}{*}{\texttt{Llama-3.2-3B}}  
    & Mild Contamination & 97.0 & 93.2 & 98.8 & 86.0 & 50.0 \\
    & Intensive Contamination & 96.7 & 90.9 & 96.8 & 92.8 & 50.0 \\ \midrule
\multirow{2}{*}{\texttt{Yi-1.5-6B}}  
    & Mild Contamination & 98.2 & 94.8 & 98.4 & 91.8 & 72.7 \\
    & Intensive Contamination & 95.8 & 88.7 & 97.9 & 94.4 & 90.9 \\ \midrule
\multirow{2}{*}{\texttt{vicuna-7b-v1.5}}  
    & Mild Contamination & 96.1 & 89.2 & 94.6 & 77.7 & 65.6 \\
    & Intensive Contamination & 96.3 & 87.2 & 93.7 & 88.5 & 90.6 \\ \midrule
\multirow{2}{*}{\texttt{Llama-3.1-8B}}  
    & Mild Contamination & 98.8 & 97.1 & 98.7 & 88.0 & 76.3 \\
    & Intensive Contamination & 98.8 & 96.5 & 99.4 & 96.2 & 94.7 \\ \midrule
\multirow{2}{*}{\texttt{Falcon3-10B}}  
    & Mild Contamination & 99.3 & 97.8 & 97.6 & 89.3 & 44.4 \\
    & Intensive Contamination & 99.6 & 98.2 & 98.3 & 91.0 & 46.3 \\ \midrule
\multirow{2}{*}{\texttt{Qwen2.5-14B}}  
    & Mild Contamination & 98.4 & 97.3 & 95.4 & 96.2 & 65.8 \\
    & Intensive Contamination & 99.9 & 98.3 & 98.6 & 97.0 & 76.3 \\ \midrule
\multirow{2}{*}{\texttt{Phi-3-medium}}  
    & Mild Contamination & 99.1 & 98.1 & 99.3 & 91.1 & 68.2 \\
    & Intensive Contamination & 99.8 & 97.5 & 99.7 & 94.0 & 68.2 \\ \midrule
\multirow{2}{*}{\texttt{DeepSeek-V2-Lite}}  
    & Mild Contamination & 97.5 & 94.5 & 97.0 & 83.2 & 37.5 \\
    & Intensive Contamination & 98.9 & 94.9 & 96.0 & 88.1 & 50.0 \\ \midrule
\multirow{2}{*}{\texttt{Qwen2.5-32B}}  
    & Mild Contamination & 99.3 & 99.3 & 97.5 & 96.5 & 61.1 \\
    & Intensive Contamination & 99.8 & 99.3 & 99.4 & 97.4 & 80.6 \\ \midrule
\multirow{2}{*}{\texttt{Yi-1.5-34B}}  
    & Mild Contamination & 99.2 & 97.8 & 99.1 & 91.4 & 89.5 \\
    & Intensive Contamination & 100.0 & 98.1 & 99.8 & 93.9 & 100.0 \\ 

\bottomrule[1.5pt]
\end{tabular}}
\end{table}

\subsubsection{Retention of General Capabilities}
\label{append:contamination validity}
A contaminated model must retain its general capabilities; otherwise, evaluation results from a severely degraded model would be meaningless. To verify this, we compute model perplexity on Alpaca~\cite{taori2023stanford}, a held-out general-purpose instruction-tuning dataset. As shown in Tab.~\ref{tab:ppl}, model perplexity remains largely unchanged after contamination, confirming that our fine-tuning process preserves general capabilities while effectively introducing benchmark contamination.

\begin{table}[h!]
\caption{Perplexity of models before and after contamination, computed on 5,000 randomly selected samples from Alpaca. ``Clean" refers to the model before contamination.}
\label{tab:ppl}
\centering
\resizebox{0.7\linewidth}{!}{
\small 
\begin{tabular}{l|c|ccccc}
\toprule[1.5pt]
\textbf{Model} & \textbf{Recipe} & \textbf{Arc-C} & \textbf{MMLU} & \textbf{TruthfulQA} & \textbf{GSM8K} & \textbf{RepliQA}\\
\midrule
\multirow{3}{*}{\texttt{Llama-3.2-3B}}  
    & Clean              & 10.83& 10.83& 10.83& 10.83& 10.83\\
    & Mild Contamination & 9.78& 9.06& 10.05& 10.43 & 10.60\\
    & Intensive Contamination & 9.96& 9.50&10.68& 13.75 & 14.57\\
\midrule
\multirow{3}{*}{\texttt{Yi-1.5-6B}}  
    & Clean              & 7.67 & 7.67& 7.67& 7.67& 7.67\\
    & Mild Contamination & 5.80& 5.57& 6.02& 6.81 & 6.48\\
    & Intensive Contamination & 6.37&6.20&6.48&  7.00 & 10.92\\
\midrule
\multirow{3}{*}{\texttt{vicuna-7b-v1.5}}  
    & Clean              & 6.87& 6.87& 6.87& 6.87& 6.87\\
    & Mild Contamination & 6.16& 5.74& 6.42& 6.91 & 6.57\\
    & Intensive Contamination & 6.34& 5.92&6.57& 7.85 & 7.56\\
\midrule
\multirow{3}{*}{\texttt{Llama-3.1-8B}}  
    & Clean              & 9.23& 9.23& 9.23& 9.23& 9.23\\
    & Mild Contamination &8.69& 8.29& 8.84& 9.83 & 9.74\\
    & Intensive Contamination &9.37& 9.17&9.57& 12.89 & 15.06\\
\midrule
\multirow{3}{*}{\texttt{Falcon3-10B}}  
    & Clean              & 7.37& 7.37& 7.37& 7.37& 7.37\\
    & Mild Contamination &4.95& 4.38& 5.11& 5.41 & 5.51\\
    & Intensive Contamination &5.70 & 4.96&5.81& 6.89 & 5.81\\
\midrule
\multirow{3}{*}{\texttt{Qwen2.5-14B}}  
    & Clean              & 5.26& 5.26& 5.26& 5.26 & 5.26\\
    & Mild Contamination & 4.93& 4.93& 5.01& 6.33 & 5.96\\
    & Intensive Contamination & 4.80& 5.08&4.96&5.72 & 4.96\\
\midrule
\multirow{3}{*}{\texttt{Phi-3-medium}}  
    & Clean              & 3.12 & 3.12& 3.12& 3.12& 3.12\\
    & Mild Contamination & 3.08& 2.67& 3.16& 3.17 & 3.08\\
    & Intensive Contamination &3.07 &2.75&3.07& 3.16 & 3.07\\
\midrule
\multirow{3}{*}{\texttt{DeepSeek-V2-Lite}}  
    & Clean              & 7.53& 7.53& 7.53& 7.53& 7.53\\
    & Mild Contamination & 6.85& 6.12& 6.75& 6.88 & 7.18\\
    & Intensive Contamination & 7.38& 6.39&7.41& 7.65 & 7.41\\
\midrule
\multirow{3}{*}{\texttt{Qwen2.5-32B}}  
    & Clean              & 4.71& 4.71& 4.71& 4.71& 4.71\\
    & Mild Contamination & 3.84& 3.85& 3.87& 4.44 & 3.94\\
    & Intensive Contamination &4.28& 4.28&4.35& 4.48 & 4.35\\
\midrule
\multirow{3}{*}{\texttt{Yi-1.5-34B}}  
    & Clean              & 6.75& 6.75& 6.75& 6.75& 6.75\\
    & Mild Contamination & 5.33& 5.09& 5.21& 8.73 & 6.31\\
    & Intensive Contamination &5.83 & 6.01&5.48& 7.10 & 5.48\\
\bottomrule[1.5pt]
\end{tabular}}
\end{table}

\subsection{Mitigation Strategy Details}

\subsubsection{Implementation and Examples} \label{Append:mitigation_examples}
We employ GPT-4o-2024-08-06 with a temperature of 0.7 to apply all mitigation strategies. We present examples showing how each of the 20 mitigation strategies updates a specific test sample from Arc-C.

\begin{itemize}
    \item Vanilla: \texttt{Which statement best describes the effect of the Sun on the oceans?\\A: The Sun influences the formation of waves.\\B: The Sun creates water particles.\\C: The Sun's rays cause organisms to come to the surface.\\D: The Sun provides minerals.\\Answer: A}
    
     \item $S_1$ Irrelevant Context: \texttt{https://t.co/DlI9kw Which statement best describes the effect of the Sun on the oceans?\\A: The Sun influences the formation of waves.\\B: The Sun creates water particles.\\C: The Sun's rays cause organisms to come to the surface.\\D: The Sun provides minerals.\\Answer: A}
     
     \item $S_2$ Relevant Context: \texttt{As the golden rays of dawn break over the horizon, the vast oceans begin to shimmer under the Sun's influence. Marine life stirs, and the water's surface reflects the Sun's warmth, bringing life to the depths below. Which statement best describes the effect of the Sun on the oceans?\\A: The Sun influences the formation of waves.\\B: The Sun creates water particles.\\C: The Sun's rays cause organisms to come to the surface.\\D: The Sun provides minerals.\\Answer: A}
     
     \item $S_3$ Syntactic Modification: \texttt{The effect of the Sun on the oceans is best described by which statement?\\A: The Sun influences the formation of waves.\\B: The Sun creates water particles.\\C: The Sun's rays cause organisms to come to the surface.\\D: The Sun provides minerals.\\Answer: A}
     
     \item $S_4$ Synonym Replacement: \texttt{Which statement best outlines the impact of the Sun on the oceans?\\A: The Sun influences the formation of waves.\\B: The Sun creates water particles.\\C: The Sun's rays cause organisms to come to the surface.\\D: The Sun provides minerals.\\Answer: A}
     
     \item $S_5$ Typographical Perturbation: \texttt{Which statemnt best descibes the efect of the Sun on the oceans?\\A: The Sun influences the formation of waves.\\B: The Sun creates water particles.\\C: The Sun's rays cause organisms to come to the surface.\\D: The Sun provides minerals.\\Answer: A}
     
     \item $S_6$ Translation (Chinese): \begin{CJK}{UTF8}{gbsn}\texttt{哪种说法最能描述太阳对海洋的影响？\\A: 太阳影响波浪的形成。\\B: 太阳产生水粒子。\\C: 太阳的光线导致生物浮到表面。\\D: 太阳提供矿物质。\\Answer: A}\end{CJK} 
     
     \item $S_7$ Translation (French): \texttt{Quelle affirmation décrit le mieux l'effet du Soleil sur les océans?\\A: Le Soleil influence la formation des vagues.\\B: Le Soleil crée des particules d'eau.\\C: Les rayons du Soleil poussent les organismes à remonter à la surface.\\D: Le Soleil fournit des minéraux.\\Answer: A}
     
     \item $S_8$ Back-translation: \texttt{Which statement best describes the Sun's effect on the ocean?\\A: The Sun influences the formation of waves.\\B: The Sun produces water particles.\\C: The Sun's rays cause organisms to float to the surface.\\D: The Sun provides minerals.\\Answer: A}
     
     \item $S_9$ Choice Paraphrasing: \texttt{Which statement best describes the Sun's effect on the ocean?\\A: The Sun affects the generation of waves.\\B: The Sun produces water particles.\\C: The sunlight encourages organisms to rise to the surface.\\D: The Sun supplies minerals.\\Answer: A}
     
     \item $S_{10}$ Additional Incorrect Choices: \texttt{Which statement best describes the Sun's effect on the ocean?\\A: The Sun influences the formation of waves.\\B: The Sun creates water particles.\\C: The Sun's rays cause organisms to come to the surface.\\D: The Sun provides minerals.\\E: The Sun alters the gravitational pull of Earth.\\F: The Sun generates tides directly.\\Answer: A}
     
     \item $S_{11}$ Choices Permutation: \texttt{Which statement best describes the Sun's effect on the ocean?\\A: The Sun creates water particles.\\B: The Sun influences the formation of waves.\\C: The Sun's rays cause organisms to come to the surface.\\D: The Sun provides minerals.\\Answer: B}
     
     \item $S_{12}$ Clean-Eval: \texttt{What is the most accurate description of how the Sun influences the ocean?\\A: The Sun influences the formation of waves.\\B: The Sun creates water particles.\\C: The Sun's rays cause organisms to come to the surface.\\D: The Sun provides minerals.\\Answer: A}
     
     \item $S_{13}$ ITD: \texttt{What is the primary influence of the Sun on oceanic conditions?\\A: The Sun affects the creation of ocean waves.\\B: The Sun generates water molecules.\\C: Sunlight causes marine life to rise to the surface.\\D: The Sun supplies nutrients.\\Answer: A}
     
     \item $S_{14}$ MPA: \texttt{The Sun, as the closest star to Earth, plays a crucial role in many natural processes. It provides light and warmth, which are essential for life on our planet. Considering its impact on various ecosystems, how does the Sun influence the behavior and characteristics of ocean waters?\\A: The Sun supplies nutrients.\\B: The Sun plays a role in creating waves.\\C: The Sun generates water molecules.\\D: The Sun's light causes living things to rise to the surface.\\E: The Sun affects the ocean's salinity levels.\\Answer: B}
 
     \item $S_{15}$ MPA-Ques+Trans-CN: \begin{CJK}{UTF8}{gbsn}\texttt{作为距离地球最近的恒星，太阳在许多自然过程中起着关键作用。它提供光和热，这对我们星球上的生命至关重要。考虑到它对各种生态系统的影响，太阳如何影响海洋水体的行为和特征？\\A: 太阳影响波浪的形成。\\B: 太阳创造水分子。\\C: 太阳的光线导致生物浮出水面。\\D: 太阳提供矿物质。\\Answer: A}\end{CJK} 
     
     \item $S_{16}$ MPA-Choice+Trans-CN: \begin{CJK}{UTF8}{gbsn}\texttt{哪种说法最能描述太阳对海洋的影响？\\A: 太阳在制造海浪中发挥作用。\\B: 太阳产生水分子。\\C: 太阳的光线使生物上升到水面。\\D: 太阳提供养分。\\E: 太阳影响海洋的盐度水平。\\Answer: A}\end{CJK}
     
     \item $S_{17}$ Mimicking: \texttt{Which statement best describes the role of the Moon on ocean tides?\\A. The Moon generates ocean currents.\\B. The Moon creates tidal waves.\\C. The Moon's gravity influences tidal movements.\\D. The Moon provides nutrients.\\Answer: C}
     
     \item $S_{18}$ Remember-Understand Extension: \texttt{What is the precise role of the Sun in driving the Earth's oceanic circulation systems?\\A. The Sun directly heats the ocean surface, causing water to evaporate and form clouds, which then drive the oceanic currents.\\B. The Sun primarily heats the ocean surface unevenly, leading to differences in water temperature that create wind patterns, which in turn drive ocean currents.\\C. The Sun's gravitational pull directly influences the tides, which are the main drivers of oceanic circulation.\\D. The Sun heats the ocean uniformly, resulting in a stable, non-dynamic oceanic system.\\Answer: B}
     
     \item $S_{19}$ Application Extension: \texttt{Considering the Sun's role in driving oceanic processes, how would you model the impact of solar radiation on the formation and movement of ocean currents?\\A. Solar radiation heats the ocean surface, causing changes in water density that drive the global thermohaline circulation.\\B. Solar radiation primarily affects only the surface temperature, having minimal impact on deeper ocean currents.\\C. The Sun's influence is limited to tidal forces, which are primarily responsible for ocean current formation.\\D. Solar radiation directly controls the salinity of the oceans, which is the sole driver of ocean currents.\\Answer: A}
     
     \item $S_{20}$ Analysis Extension: \texttt{Analyze the influence of solar radiation on oceanic currents and temperature gradients. How do these factors interact to affect global climate patterns?\\A. Solar radiation directly heats the ocean surface, creating horizontal temperature gradients that drive ocean currents, which in turn redistribute heat and influence global climate.\\B. Solar radiation causes vertical temperature gradients by heating the ocean depths, resulting in convection currents that primarily affect local weather patterns.\\C. The Sun's influence on the ocean is minimal, with most oceanic currents and temperature gradients driven by geothermal activity rather than solar heating.\\D. Solar radiation causes uniform heating of the ocean surface, leading to a stable temperature across all latitudes and minimal impact on global climate patterns.\\Answer: A}

\end{itemize}

\subsubsection{Updated data with wrong answers} 
\label{Append:wrong_mitigation_examples}
We include a qualitative example of incorrect answers generated by LLMs due to limitations in their domain-specific knowledge. We provide an MMLU example where human expert evaluation indicates that the answer generated by the Remember-Understand Extension strategy is incorrect.

\begin{itemize}
    \item Vanilla: \texttt{For which of these two scenarios does the main character (who uses I/me/my) do something clearly morally wrong, according to ordinary moral standards in the US as of 2020? Scenario 1 | I left the hotel room's clothes iron on since I expected that the housekeepers would eventually take care of it. Scenario 2 | In my midlife crisis, I abandoned my family.\\A: Wrong, Wrong\\B. Wrong, Not wrong\\C. Not wrong, Wrong\\D. Not wrong, Not wrong\\Answer: A}
    \item Remember-Understand Extension (Fidelity: 0.655): \texttt{What is the philosophical definition of `moral wrong' according to contemporary ethics?\\A. An action is morally wrong if it violates a universal law applicable to everyone without exception.\\B. An action is morally wrong only if it causes harm to the person performing the action.\\C. An action is morally wrong if it is against the subjective preferences of the majority population.\\D. An action is morally wrong if it goes against the principles of fairness and justice universally recognized in a society.\\LLM-generated Answer: A.\\ \textcolor{red}{Correct answer: D.}}
\end{itemize}

\subsection{LLM Evaluation Details}
The evaluation template follows the format:
\texttt{Question:\{input\}\textbackslash n Answer:},
where ``input'' includes the question and choices (if applicable).
For multiple-choice benchmarks (Arc-C, MMLU, and TruthfulQA), we adopt a zero-shot evaluation approach, selecting the option with the highest probability as the predicted answer. We also conduct an ablation study using a 25-shot evaluation on Arc-C. The results remain consistent with our primary conclusions.

For GSM8K, we use a 5-shot evaluation approach with a specific prompt shown below, setting the maximum number of generated tokens to 256. The numerical answer is extracted using regex by matching the digits following the ``\#\#\#\#'' symbol.

For RepliQA, we employ a zero-shot evaluation approach with a maximum generation length of 128 tokens. The generated answers are evaluated by GPT-4o-mini, which compares the predicted answer with the ground truth and assigns a binary correctness score (0 for incorrect, 1 for correct).

\begin{center}
\begin{tcolorbox}[title={The 5-shot prompt used for GSM8K evaluation. }]
\texttt{
Question: Jen and Tyler are gymnasts practicing flips. Jen is practicing the triple-flip while Tyler is practicing the double-flip. Jen did sixteen triple-flips during practice. Tyler flipped in the air half the number of times Jen did. How many double-flips did Tyler do?\textbackslash n Answer: Jen did 16 triple-flips, so she did 16 * 3 = <<16*3=48>>48 flips.\textbackslash n Tyler did half the number of flips, so he did 48 / 2 = <<48/2=24>>24 flips.\textbackslash n A double flip has two flips, so Tyler did 24 / 2 = <<24/2=12>>12 double-flips.\textbackslash n\#\#\#\# 12\textbackslash n\textbackslash n Question: Four people in a law firm are planning a party. Mary will buy a platter of pasta for \$20 and a loaf of bread for \$2. Elle and Andrea will split the cost for buying 4 cans of soda which cost \$1.50 each, and chicken wings for \$10. Joe will buy a cake that costs \$5. How much more will Mary spend than the rest of the firm put together?\textbackslash n Answer: Mary will spend \$20 + \$2 = \$<<20+2=22>>22.\textbackslash n Elle and Andrea will spend \$1.5 x 4 = \$<<1.5*4=6>>6 for the soda.\textbackslash n Elle and Andrea will spend \$6 + \$10 = \$<<6+10=16>>16 for the soda and chicken wings.\textbackslash n Elle, Andrea, and Joe together will spend \$16 + \$5 = \$<<16+5=21>>21.\textbackslash n So, Mary will spend \$22 - \$21 = \$<<22-21=1>>1 more than all of them combined.\textbackslash n\#\#\#\# 1\textbackslash n\textbackslash n Question: A charcoal grill burns fifteen coals to ash every twenty minutes of grilling. The grill ran for long enough to burn three bags of coals. Each bag of coal contains 60 coals. How long did the grill run?\textbackslash n Answer: The grill burned 3 * 60 = <<3*60=180>>180 coals.\textbackslash n It takes 20 minutes to burn 15 coals, so the grill ran for 180 / 15 * 20 = <<180/15*20=240>>240 minutes.\textbackslash n\#\#\#\# 240\textbackslash n\textbackslash n Question: A bear is preparing to hibernate for the winter and needs to gain 1000 pounds. At the end of summer, the bear feasts on berries and small woodland animals. During autumn, it devours acorns and salmon. It gained a fifth of the weight it needed from berries during summer, and during autumn, it gained twice that amount from acorns. Salmon made up half of the remaining weight it had needed to gain. How many pounds did it gain eating small animals?\textbackslash n Answer: The bear gained 1 / 5 * 1000 = <<1/5*1000=200>>200 pounds from berries.\textbackslash n It gained 2 * 200 = <<2*200=400>>400 pounds from acorns.\textbackslash n It still needed 1000 - 200 - 400 = <<1000-200-400=400>>400 pounds.\textbackslash n Thus, it gained 400 / 2 = <<400/2=200>>200 pounds from salmon.\textbackslash n Therefore, the bear gained 400 - 200 = <<400-200=200>>200 pounds from small animals.\textbackslash n\#\#\#\# 200\textbackslash n\textbackslash n Question: Brendan can cut 8 yards of grass per day, he bought a lawnmower and it helped him to cut more yards by Fifty percent per day. How many yards will Brendan be able to cut after a week?\textbackslash n Answer: The additional yard Brendan can cut after buying the lawnmower is 8 x 0.50 = <<8*0.50=4>>4 yards.\textbackslash n So, the total yards he can cut with the lawnmower is 8 + 4 = <<8+4=12>>12.\textbackslash n Therefore, the total number of yards he can cut in a week is 12 x 7 = <<12*7=84>>84 yards.\textbackslash n\#\#\#\# 84\textbackslash n
}
\label{gsm8k_prompt}
\end{tcolorbox}
\end{center}
